\RequirePackage[l2tabu, orthodox]{nag}
\documentclass[11pt]{article}

\usepackage[T1]{fontenc}

\usepackage[bitstream-charter]{mathdesign}
\usepackage{amsmath}
\usepackage[scaled=0.92]{PTSans}

\usepackage[
  paper  = letterpaper,
  left   = 1.65in,
  right  = 1.65in,
  top    = 1.0in,
  bottom = 1.0in,
  ]{geometry}

\usepackage[usenames,dvipsnames]{xcolor}
\definecolor{shadecolor}{gray}{0.9}

\usepackage[final,expansion=alltext]{microtype}
\usepackage[english]{babel}
\usepackage[parfill]{parskip}
\usepackage{afterpage}
\usepackage{framed}

\DeclareRobustCommand{\parhead}[1]{\textbf{#1}~}

\usepackage{lineno}

\usepackage{ragged2e}

\newcounter{parcount}

\usepackage{graphicx}
\usepackage[labelfont=bf]{caption}

\usepackage{booktabs}
\usepackage{natbib}
\usepackage[algoruled]{algorithm2e}
\usepackage{listings}
\usepackage{fancyvrb}
\fvset{fontsize=\normalsize}

\usepackage[colorlinks,linktoc=all]{hyperref}
\usepackage[all]{hypcap}
\hypersetup{citecolor=MidnightBlue}
\hypersetup{linkcolor=black}
\hypersetup{urlcolor=MidnightBlue}

\usepackage[acronym,smallcaps,nowarn]{glossaries}

\lstdefinestyle{mystyle}{
    commentstyle=\color{OliveGreen},
    keywordstyle=\color{BurntOrange},
    numberstyle=\tiny\color{black!60},
    stringstyle=\color{MidnightBlue},
    basicstyle=\ttfamily,
    breakatwhitespace=false,
    breaklines=true,
    captionpos=b,
    keepspaces=true,
    numbers=left,
    numbersep=5pt,
    showspaces=false,
    showstringspaces=false,
    showtabs=false,
    tabsize=2
}
\DeclareMathOperator*{\argmax}{arg\,max}
\DeclareMathOperator*{\argmin}{arg\,min}

\newcommand{\g}{\, | \,}

\newcommand{\bz}{\bm{z}}
\newcommand{\bx}{\bm{x}}

\newcommand{\bw}{\bm{w}}

\newcommand{\bv}{\bm{v}}

\newcommand{\bmu}{\bm{\mu}}
\newcommand{\btheta}{\bm{\theta}}
\newcommand{\bbeta}{\bm{\beta}}

\newcommand{\balpha}{\bm{\alpha}}
\newcommand{\bSigma}{\bm{\Sigma}}

 \newacronym{ALI}{ali}{adversarially learned inference}
\newacronym{BIGAN}{bigan}{bidirectional generative adversarial network}
\newacronym{VI}{vi}{variational inference}
\newacronym{KL}{kl}{Kullback-Leibler}
\newacronym{ELBO}{elbo}{evidence lower bound}
\newacronym{MCMC}{mcmc}{Markov chain Monte Carlo}
\newacronym{HMC}{hmc}{Hamiltonian Monte Carlo}
\newacronym{RNN}{rnn}{recurrent neural network}
\newacronym{MLP}{mlp}{feed forward neural network}
\newacronym{DEF}{def}{deep exponential family}
\newacronym{LSTM}{lstm}{long-short term memory}
\newacronym{GRU}{gru}{gated recurrent unit}
\newacronym{VRNN}{vrnn}{variational recurrent neural network}
\newacronym{SRNN}{srnn}{stochastic recurrent neural network}
\newacronym{ERNN}{ernn}{Elman recurrent neural network}
\newacronym{VAE}{vae}{variational auto-encoder}
\newacronym{DCVAE}{dcvae}{deep-convolutional variational auto-encoder}
\newacronym{UIVI}{uivi}{unbiased implicit variational inference}
\newacronym{DGM}{dgm}{deep generative model}
\newacronym{IWAE}{iwae}{importance weighted auto-encoder}
\newacronym{RWS}{rws}{reweighted wake-sleep}
\newacronym{REM}{rem}{reweighted expectation maximization}
\newacronym{EM}{em}{expectation maximization} 
\usepackage{subfigure}
\usepackage{bm}
\usepackage{authblk}

\title{\textbf{Reweighted Expectation Maximization}}

\author[1]{Adji B. Dieng}
\author[2]{John Paisley}
\affil[1]{Department of Statistics, Columbia University}
\affil[2]{Department of Electrical Engineering, Columbia University}

\date{}

\begin{document}
\maketitle

\begin{abstract}\noindent Training deep generative models with maximum likelihood remains a challenge. The typical workaround is to use \gls{VI} and maximize a lower bound to the log marginal likelihood of the data. Variational auto-encoders (\acrshort{VAE}s) adopt this approach. They further amortize the cost of inference by using a recognition network to parameterize the variational family. Amortized \gls{VI}  scales approximate posterior inference in deep generative models to large datasets. However it introduces an \emph{amortization gap} and leads to approximate posteriors of reduced expressivity due to the problem known as \emph{posterior collapse}. 
In this paper, we consider \gls{EM} as a paradigm for fitting deep generative models. Unlike \gls{VI}, \gls{EM} directly maximizes the log marginal likelihood of the data. We rediscover the \gls{IWAE} as an instance of \gls{EM} and propose a new \gls{EM}-based algorithm for fitting deep generative models called \gls{REM}. \gls{REM} learns better generative models than the \gls{IWAE} by decoupling the learning dynamics of the generative model and the recognition network using a separate expressive proposal found by moment matching. We compared \gls{REM} to the \acrshort{VAE} and the \gls{IWAE} on several density estimation benchmarks and found it leads to significantly better performance as measured by log-likelihood\footnote{\textbf{Code:} Code for this work an be found at \url{https://github.com/adjidieng/REM}}.\\

\end{abstract}

\section{Introduction}
\glsresetall

Parameterizing latent variable models with deep neural networks is becoming a major approach to probabilistic modeling~\citep{hinton2006fast, salakhutdinov2009deep, gregor2013deep, kingma2013auto, rezende2014stochastic}. These models are very expressive. However, challenges arise when learning the posterior distribution of the latent variables and the model parameters. One main inference technique is \gls{VI}~\citep{jordan1999introduction, blei2017variational}. It consists in choosing a variational distribution to approximate the true posterior and then finding the parameters of the variational distribution that maximize the \gls{ELBO}, a lower bound on the log marginal likelihood of the data. In deep latent variable models, the variational distribution is parameterized by a recognition network---a deep neural network that takes data as input and outputs the parameters of a distribution~\citep{dayan1995helmholtz, kingma2013auto, rezende2014stochastic}. The model and recognition network parameters are learned jointly by maximizing the \gls{ELBO}. 

Approximating the true posterior using a recognition network and maximizing the \gls{ELBO} enables efficient learning in large data settings. However this procedure introduces an \emph{amortization gap}~\citep{cremer2018inference}, and leads to learned approximate posteriors that may lack expressivity due to the ``posterior collapse" problem~\citep{bowman2015generating, hoffman2016elbo, sonderby2016train, kingma2016improved, chen2016variational, dieng2018avoiding, he2019lagging, razavi2019preventing}. 

Several other algorithms have been proposed to fit deep generative models (e.g.~\citet{bornschein2014reweighted, burda2015importance, rezende2015variational, kingma2016improved}.) Some are based on importance sampling, in which several samples are drawn from the approximate posterior \citep{bornschein2014reweighted, burda2015importance}.

In this paper, we propose returning to \gls{EM} as an alternative to variational inference for fitting deep generative models.  
\gls{EM} has been originally applied to problems where one aims to perform maximum likelihood in the presence of missing data~\citep{dempster1977maximum}. It has since been used in other problems, for example in reinforcement learning~\citep{dayan1997using}. As opposed to traditional variational inference, which maximizes a lower bound to the log marginal likelihood of the data, \gls{EM} directly targets the log marginal likelihood. Each iteration in \gls{EM} is guaranteed to increase the log marginal likelihood from the previous iteration~\citep{bishop2006pattern}. 

Using \gls{EM} in the context of deep generative models should lead to better generative models. In fact we show that the \gls{IWAE}~\citep{burda2015importance}, which achieves better performance in density estimation than the \gls{VAE}~\citep{kingma2013auto, rezende2014stochastic}, is an instance of \gls{EM}. 

We take advantage of this observation to propose an algorithm called \gls{REM} that improves upon the \gls{IWAE} (and the \gls{VAE}) on density estimation. \gls{REM} decouples the learning dynamics of the generative model and the recognition network using an expressive proposal found by moment matching. 
This decoupling prevents the generative model from co-adapting with the recognition network, a problem that the \gls{VAE} is known to suffer from~\citep{cremer2018inference}. 

We compared \gls{REM} against the \gls{VAE} and the \gls{IWAE} on several density estimation benchmarks. We found \gls{REM} leads to significantly better performance as measured by  log-likelihood. 

The rest of the paper is organized as follows. In Section\nobreakspace \ref {sec:related} we discuss some related work. In Section\nobreakspace \ref {sec:background} we review \gls{VI} and \gls{EM} and emphasize their differences. In Section\nobreakspace \ref {sec:method} we propose \gls{EM} as an inference method for deep generative models which leads us to rediscover the \gls{IWAE} and propose a new inference algorithm for fitting deep generative models called \gls{REM}. We then compare the performance of \gls{REM} against the \gls{VAE} and the \gls{IWAE} in Section\nobreakspace \ref {sec:empirical}. Finally, we conclude in Section\nobreakspace \ref {sec:discussion}.

\section{Related Work}\label{sec:related}

Deep generative modeling is an approach to unsupervised representation learning that has shown great promise~\citep{kingma2013auto, rezende2014stochastic, goodfellow2014generative, dinh2016density}. Early deep generative models include belief networks~\citep{neal1992connectionist, hinton2009deep}, the Hemholtz machine~\citep{hinton1995wake}, and the deep Boltzmann machine~\citep{salakhutdinov2009deep}. More recently \citet{kingma2013auto, rezende2014stochastic} proposed the \gls{VAE}.

\acrshort{VAE}s are the result of combining variational Bayesian methods with the flexibility and scalability of neural networks~\citep{kingma2013auto, rezende2014stochastic}, and have been used in various applications (e.g.~\citet{bowman2015generating, gregor2015draw, zhao2018unsupervised, liang2018variational}). 
However \glspl{VAE} are notoriously known to suffer from a problem called latent variable collapse discussed in several works~\citep{bowman2015generating, hoffman2016elbo, sonderby2016train, kingma2016improved, chen2016variational, alemi2017fixing, higgins2017beta, dieng2018avoiding, he2019lagging}. As a result of latent variable collapse the learned latent representations are overly simplified and poorly represent the underlying structure of the data. 

The \gls{IWAE} was introduced to prevent posterior collapse and learn better generative models~\citep{burda2015importance}. The \gls{IWAE} relies on importance sampling to optimize both the model parameters and the recognition network. The \gls{IWAE} objective is shown to be a tighter lower bound of the log marginal likelihood of the data than the \gls{ELBO}~\citep{burda2015importance}. The tightness of the bound is determined by the number of particles used for importance sampling. It has been shown that increasing the number of particles leads to poorer recognition networks due to a diminishing signal-to-noise ratio in the gradients of the \gls{IWAE} objective~\citep{rainforth2018tighter, le2018revisiting}. \citet{le2018revisiting} also show that the \gls{RWS}~\citep{bornschein2014reweighted} does not suffer from this issue. The \gls{RWS} extends the wake-sleep algorithm~\citep{hinton1995wake} to importance sampling the same way the \gls{IWAE} extends the \gls{VAE} to importance sampling. 

Most of the algorithms discussed above use a variational inference perspective to fit generative models with an \gls{EM} objective as the starting point. We propose directly using the \gls{EM} perspective as an alternative. \gls{EM} was first introduced in the statistics literature, where it was used to solve problems involving missing data~\citep{dempster1977maximum}. One typical application of the \gls{EM} algorithm is to fit mixtures of Gaussians, where the cluster assignments are considered \emph{unobserved} data~\citep{bishop2006pattern, murphy2012machine}. Other applications of \gls{EM} arise in conjugate graphical models. (See \citet{murphy2012machine} for examples of conjugate models using \gls{EM}.) \gls{EM} has also been applied to reinforcement learning~\citep{dayan1997using}. More recently \citet{song2016learning} used \gls{EM} to fit sigmoid belief networks~\citep{song2016learning}. In this paper we develop a general \gls{EM} procedure for fitting deep generative models.

\section{Notation and Background}
\label{sec:background}
In this section we first describe notation and nomenclature and then review variational inference and expectation maximization. In particular, we review how \gls{EM} guarantees the maximization of the log marginal likelihood after each of its iterations. 

\subsection{Notation}\label{sub:notation}
Throughout the paper we consider a set of $N$ i.i.d datapoints $\bx_1, \dots, \bx_N$. We posit each observation $\bx_i$ is drawn by first sampling a latent variable $\bz_i$ from some fixed prior $p(\bz)$ and then sampling $\bx_i$ from $p_{\theta}(\bx_i \g \bz_i)$---the conditional distribution of $\bx_i$ given $\bz_i$. We parameterize the conditional $p_{\theta}(\bx_i \g \bz_i)$ using a deep neural network and $\theta$ represents the parameters of this network and any other parameters used to define the model. Our goal is to learn the parameters $\theta$ and the posterior distribution of the latents given the observations, $p_{\theta}(\bz_i \g \bx_i)$. We denote by \emph{proposal} any auxiliary distribution involved in the learning of the parameters $\theta$. We call \emph{hyperobjective} and \emph{hyperproposal} any auxiliary objective and distribution used to learn the proposal, respectively.

\subsection{Variational Inference}
Variational inference (\gls{VI}) is a scalable approach to approximate posterior inference. It first assumes a family of distributions over the latent variables and then finds the member of this family that best approximates the true posterior. The quality of the approximation is measured by how close the approximate posterior is to the true posterior. Closeness is determined by a divergence measure; typically the reverse \gls{KL} divergence. Minimizing this divergence is intractable as it still depends on the unknown true posterior. The approach in \gls{VI} is to instead maximize a lower bound to the log marginal likelihood of the data. 

More specifically consider the same set up as in Section\nobreakspace \ref {sub:notation} but focus on one observation denoted by $\bx$ for simplicity. 
Bayes rule writes the true posterior distribution of the latent $\bz$ given $\bx$ as a function of the prior and the likelihood, $p_{\theta}(\bz \g \bx) \propto p_{\theta}(\bx \g \bz) \cdot p(\bz)$. \gls{VI} approximates this posterior distribution using a variational distribution $q_{\phi}(\bz)$ whose parameters $\phi$ are learned jointly with the model parameters $\theta$ by maximizing the \gls{ELBO},
\begin{align}\label{eq:elbo}
	\gls{ELBO} &=  \mathbb{E}_{q_{\phi}(\bz)}\left[ \log p_{\theta}(\bx , \bz) - \log q_{\phi}(\bz) \right] = \mathbb{E}_{q_{\phi}(\bz)}\left[ \log p_{\theta}(\bx \g \bz)\right] -\gls{KL}(q_{\phi}(\bz) \vert\vert p(\bz))
	.
\end{align} 

\gls{VI} has been the method of choice for fitting deep generative models. In these settings, the approximate posterior $q_{\phi}(\bz)$ explicitly conditions on $\bx$ and we write $q_{\phi}(\bz \g \bx)$. It is typically a Gaussian parameterized by a recognition network that takes $\bx$ as input. This is the approach of \glspl{VAE} where the conditional $p_{\theta}(\bx \g \bz)$ is a deep neural network that takes $\bz$ as input. 
Maximizing the \gls{ELBO} in these settings enables scalable approximate posterior inference since the variational parameters are shared across all observations through a neural network, but leads to the problem known as posterior collapse. This problem occurs because the \gls{KL} term in Equation\nobreakspace \textup {(\ref {eq:elbo})} decays rapidly to zero during optimization,  leaving $q_{\phi}(\bz \g \bx)$ not representative of the data. The generative model is unable to correct this behavior as it tends to co-adapt with the choice of $q_{\phi}(\bz \g \bx)$~\citep{cremer2018inference}. 

\subsection{Expectation Maximization}
\gls{EM} is a maximum likelihood iterative optimization technique that directly targets the log marginal likelihood and served as the departure point for the development of variational inference methods. The \gls{EM} objective is the log marginal likelihood of the data,
\begin{align}\label{eq:log_marginal}
	\log p_{\theta}(\bx) &= \mathbb{E}_{q_{\phi}(\bz)}\left[ \log p_{\theta}(\bx , \bz) - \log q_{\phi}(\bz) \right] + \gls{KL}\left(q_{\phi}(\bz) \vert\vert p_{\theta}(\bz \g \bx)\right). 
\end{align}
\gls{EM} alternates between an \emph{E-step}, which sets the second term in Equation\nobreakspace \textup {(\ref {eq:log_marginal})} to zero, and an \emph{M-step}, which fits the model parameters $\theta$ by maximizing the first term using the proposal learned in the E-step. Note that after the E-step, the objective in Equation\nobreakspace \textup {(\ref {eq:log_marginal})} says the log marginal is exactly equal to the \gls{ELBO} which is a tractable objective for fitting the model parameters. \gls{EM} alternates these two steps until convergence to an approximate maximum likelihood solution for $p_{\theta}(\bx)$.

Contrast this with \gls{VI}. The true objective for \gls{VI} is the \gls{KL} term in Equation\nobreakspace \textup {(\ref {eq:log_marginal})}, $\gls{KL}\left(q_{\phi}(\bz) \vert\vert p_{\theta}(\bz \g \bx)\right)$, which is intractable. The argument in \gls{VI} is then to say that minimizing this \gls{KL} is equivalent to maximizing the \gls{ELBO}, the first term in Equation\nobreakspace \textup {(\ref {eq:log_marginal})}. This argument only holds when the log marginal likelihood $\log p_{\theta}(\bx)$ has no free parameters, in which case it is called the \emph{model evidence}. 
Importantly, \gls{VI} does not necessarily maximize $\log p_{\theta}(\bx)$ because it chooses approximate posteriors $q_{\phi}(\bz)$ that may be far from the exact conditional posterior.  

In contrast \gls{EM} effectively maximizes $\log p_{\theta}(\bx)$ after each iteration. 
Consider given $\theta_t$, the state of the model parameters after the $t^{th}$ iteration of \gls{EM}. \gls{EM} learns $\theta_{t+1}$ through two steps, which we briefly review:
\begin{align}
	\text{E-step:} &  \text{ set }  q_{\phi}(\bz) = p_{\theta_t}(\bz \g \bx)\label{eq:e_step}\\
	\text{M-step:} & \text{ define } \theta_{t+1} = \argmax_{\theta} \mathcal{L}(\theta) = \mathbb{E}_{q_{\phi}(\bz)}\left[ \log p_{\theta}(\bx , \bz) - \log q_{\phi}(\bz) \right] \label{eq:m_step}
\end{align}
The value of the log marginal likelihood for $\theta_{t+1}$ is greater than for  $\theta_t$. To see this, write
\begin{align*}
	\log p_{\theta_t}(\bx) &= \mathcal{L}(\theta_t)  + \gls{KL}\left(q_{\phi}(\bz) \vert\vert p_{\theta_t}(\bz \g \bx)\right) = \mathcal{L}(\theta_t) \\
	&\leq \mathcal{L}(\theta_{t+1}) 
	\leq \mathcal{L}(\theta_{t+1}) + \gls{KL}\left(q_{\phi}(\bz) \vert\vert p_{\theta_{t+1}}(\bz \g \bx)\right) = \log p_{\theta_{t+1}}(\bx)
\end{align*}
where the second equality is due to the E-step, the first inequality is due to the M-step, and the second inequality is due to the nonnegativity of \gls{KL}.

\section{Reweighted Expectation Maximization}
\label{sec:method}
We consider \gls{EM} as a paradigm for fitting deep generative models. We are in the modeling regime where there are $N$ iid datapoints $(\bx_1, \dots, \bx_N)$ and a latent variable $\bz_i$ for each datapoint $\bx_i$, and use the same notation as Section\nobreakspace \ref {sec:background}. Assume given $\theta_t$ from the previous iteration of \gls{EM} and consider the E-step in Equation\nobreakspace \textup {(\ref {eq:e_step})}. Factorize the proposal $q_{\phi}(\bz)$ the same way as the true posterior factorizes, that is
\begin{align}\label{eq:factorization}
	\prod_{i=1}^{N} q_{\phi}(\bz_i) &= \prod_{i=1}^{N} p_{\theta_t}(\bz_i \g \bx_i).
\end{align}
The equality in Equation\nobreakspace \textup {(\ref {eq:factorization})} is achieved by setting $q_{\phi}(\bz_i)  =  p_{\theta_t}(\bz_i \g \bx_i)$ $\forall$ $i$. 
Now consider the M-step in Equation\nobreakspace \textup {(\ref {eq:m_step})}. Its goal is to find the best parameters $\theta_{t+1}$ at iteration $t+1$ that maximize  
\begin{align}\label{eq:factorization_2}
	\mathcal{L}(\theta) &=\sum_{i=1}^{N} \mathbb{E}_{p_{\theta_t}(\bz_i \g \bx_i)}\left[
		\log p_{\theta}(\bx_i, \bz_i) - \log p_{\theta_t}(\bz_i \g \bx_i)
	\right]
\end{align}
where we replaced $q_{\phi}(\bz_i)$  by $p_{\theta_t}(\bz_i \g \bx_i)$ using the E-step and wrote $\mathcal{L}(\theta)$ as a summation over the data using the model's factorization (Equation\nobreakspace \textup {(\ref {eq:factorization})}). The term $\log p_{\theta_t}(\bz_i \g \bx_i)$ is a constant with respect to $\theta$ and we can ignore it,
\begin{align}\label{eq:factorization_3}
	\mathcal{L}(\theta) 
	&=\sum_{i=1}^{N} \mathbb{E}_{p_{\theta_t}(\bz_i \g \bx_i)}\left[
		\log p_{\theta}(\bx_i, \bz_i) \right]
	= \sum_{i=1}^{N} \int_{}^{} \frac{p_{\theta_t}(\bz_i , \bx_i)}{p_{\theta_t}(\bx_i)} \log p_{\theta}(\bx_i, \bz_i) \text{ } d\bz_i
	.
\end{align}
This objective is intractable because it involves the marginal $p_{\theta_t}(\bx_i)$\footnote{Although the marginal here does not depend on $\theta$, it cannot be ignored because it depends on the $i^{th}$ datapoint. Therefore it cannot be pulled outside the summation. }. However we can make it tractable using self-normalized importance sampling~\citep{Owen2013},
\begin{align}\label{eq:rem_new}
	\mathcal{L}(\theta) 
	&= \sum_{i=1}^{N} \mathbb{E}_{r_{\eta_t}(\bz_i \g \bx_i)}\left[
		\frac{\bw(\bx_i, \bz_i; \theta_t, \eta_t) \log p_{\theta}(\bx_i, \bz_i)}{\mathbb{E}_{r_{\eta_t}(\bz_i \g \bx_i)}\left(\bw(\bx_i, \bz_i; \theta_t, \eta_t)\right)} 
	\right]
	.
\end{align}
where $\bw(\bx_i, \bz_i; \theta_t, \eta_t) = \frac{p_{\theta_t}(\bz_i , \bx_i)}{r_{\eta_t}(\bz_i \g \bx_i)}$. 
Here $r_{\eta_t}(\bz_i \g \bx_i)$ is a proposal distribution. 
Its parameter $\eta_t$ was fitted in the $t^{th}$ iteration. We now approximate the expectations in Equation\nobreakspace \textup {(\ref {eq:rem_new})} using Monte Carlo by drawing $K$ samples $\bz_i^{(1)}, \dots, \bz_i^{(K)}$ from the proposal, 
\begin{align}\label{eq:rem_samples}
\balpha_{it}^{k} = \frac{\bw(\bx_i, \bz_i^{(k)}; \theta_t, \eta_t)}{\sum_{k=1}^{K} \bw(\bx_i, \bz_i^{(k)}; \theta_t, \eta_t)} \quad \text{and} \quad
	\mathcal{L}(\theta) 
	&= \sum_{i=1}^{N} \sum_{k=1}^{K} \balpha_{it}^{k}\cdot \log p_{\theta}(\bx_i, \bz_i^{(k)})
\end{align} 
Note the approximation in Equation\nobreakspace \textup {(\ref {eq:rem_samples})} is biased but asymptotically unbiased. More specifically, the approximation improves as the number of particles $K$ increases. 

We use gradient-based learning which requires to compute the gradient of $\mathcal{L}(\theta)$ with respect to the model parameters $\theta$, this is 
\begin{align}\label{eq:grad_theta}
	\nabla_{\theta}\mathcal{L}(\theta) 
	&= \sum_{i=1}^{N} \sum_{k=1}^{K} \balpha_{it}^{k} \cdot \nabla_{\theta}\log p_{\theta}(\bx_i, \bz_i^{(k)}). 
\end{align}
The expression of the gradient in Equation\nobreakspace \textup {(\ref {eq:grad_theta})} is the same as in \gls{IWAE}~\citep{burda2015importance}. 
\gls{IWAE} was derived in \citet{burda2015importance} from the point of view of maximizing a tighter lower bound to the log marginal likelihood using importance sampling.  
Here, we have derived the \gls{IWAE} update rule for the model parameters $\theta$ using the \gls{EM} algorithm. 
The remaining question is how to define and fit the proposal $r_{\eta_t}(\bz_i \g \bx_i)$.

\begin{algorithm}[t]
  \small
  \SetAlgoNoLine
     \DontPrintSemicolon
     \SetKwInOut{KwInput}{input}
     \SetKwInOut{KwOutput}{output}
     \KwInput{Data $\bx$}
     Initialize model and proposal parameters $\theta, \eta$\;
     \For{\emph{iteration} $t=1,2,\ldots$}{
       Draw minibatch of observations $\{\bx_n\}_{n=1}^{B}$\;
        \For{\emph{observation} $n=1,2,\ldots, B$}{
       Draw $\bz^{(1)}_n, \dots, \bz^{(K)}_n \sim r_{\eta_t}(\bz_n^{(k)} \g \bx_n)$ \; 
       Compute importance weights $\bw^{(k)} = \frac{p_{\theta_t}(\bz_n^{(k)} , \bx_n)}{r_{\eta_t}(\bz_n^{(k)} \g \bx_n)}$\;        Compute $\bmu_{nt} = \sum_{k=1}^{K} \frac{\bw^{(k)}}{\sum_{k=1}^{K} \bw^{(k)}} \bz^{(k)}_n$ and $\bSigma_{nt} = \sum_{k=1}^{K} \frac{\bw^{(k)}}{\sum_{k=1}^{K} \bw^{(k)}}  (\bz^{(k)}_n - \bmu_{nt})(\bz^{(k)}_n  - \bmu_n)^\top$\;
       Set proposal $s(\bz_n^{(t)}) = \mathcal{N}(\bmu_{nt}, \bSigma_{nt})$\;
       }
       Compute $\nabla_{\eta}\mathcal{L}(\eta) = \frac{1}{\vert B\vert}\sum_{n \in B}^{} \sum_{k=1}^{K} \frac{\bv^{(k)}}{\sum_{k=1}^{K} \bv^{(k)}} \nabla_{\eta}\log r_{\eta}(\bz_n^{(k)} \g \bx_n)$ and update $\eta$ using Adam\;
       Compute $\nabla_{\theta}\mathcal{L}(\btheta)  = \frac{1}{\vert B\vert}\sum_{n \in B}^{} \sum_{k=1}^{K} \frac{\bw^{(k)}}{\sum_{k=1}^{K} \bw^{(k)}} \nabla_{\theta}\log p_{\theta}(\bx_n, \bz^{(k)}_n)$ and update $\theta$ using Adam\;
      }
     \caption{Learning with reweighted expectation maximization (\acrshort{REM} (v1))\label{alg:rem}}
     \vspace{-2pt}
\end{algorithm}

\subsection{The \gls{IWAE} proposal} 
The \gls{IWAE} uses a recognition network---a neural network that takes data $\bx_i$ as input and outputs the parameters of a distribution---as a proposal. In the \gls{IWAE}, this distribution is a diagonal Gaussian. The \gls{IWAE} fits the proposal parameters $\eta$ for the next iteration jointly with the model parameters $\theta$ using stochastic optimization. The objective for $\eta$ in the \gls{IWAE} is\footnote{This objective is to be maximized with respect to $\eta$. } 
\begin{align}
	\mathcal{L}_{\gls{IWAE}}(\eta) &= \sum_{i=1}^{N} \log \left(\frac{1}{K}\sum_{k=1}^{K} \frac{p_{\theta}(\bx_i, \bz_i^{(k)})}{r_{\eta}(\bz_i^{(k)} \g \bx_i)}\right)
.
\end{align}
As pointed out in \citet{le2017auto} this does not correspond to minimizing any divergence between the \gls{IWAE}'s proposal and the true posterior 
and leads to poor approximate posteriors as the number of samples $K$ increases~\citep{rainforth2018tighter}. We also observe this in Section\nobreakspace \ref {sec:empirical}. 

\subsection{Finding rich proposals via moment matching}
We now propose better methods for fitting the proposal. 

\parhead{Moment matching as a hyperproposal}. Denote by $\eta_t$ the proposal parameters at the previous iteration. We learn $\eta_{t+1}$ by targeting the true posterior $p_{\theta_t}(\bz \g \bx)$, 
\begin{align}\label{eq:inclusive_kl}
	\eta_{t+1} &= \argmin_{\eta} \mathcal{L}_{\gls{REM}}(\eta) = \gls{KL}(p_{\theta_t}(\bz \g \bx) \vert\vert r_{\eta}(\bz \g \bx))
	.
\end{align}
Unlike the \gls{IWAE}, the proposal here targets the true posterior using a well defined objective---the inclusive \gls{KL} divergence. The inclusive \gls{KL} induces overdispersed proposals which are beneficial in importance sampling~\citep{minka2005divergence}. 

The objective in Equation\nobreakspace \textup {(\ref {eq:inclusive_kl})} is still intractable as it involves the true posterior $p_{\theta_t}(\bz\g \bx)$, 
\begin{align}\label{eq:proposal_loss}
	\mathcal{L}_{\gls{REM}}(\eta) &= -\sum_{i=1}^{N} \mathbb{E}_{p_{\theta_t}(\bz_i \g \bx_i)}\left[\log r_{\eta}(\bz_i \g \bx_i) \right] + \text{const.},
\end{align}
where $\text{const.}$ is a constant with respect to $\eta$ that we can ignore. We use the same approach as for fitting the model parameters $\theta$. That is, we write
\begin{align}\label{eq:rem}
	\mathcal{L}_{\gls{REM}}(\eta) 
	&= \!-\!\sum_{i=1}^{N} \mathbb{E}_{s(\bz_i)}\!\left[
		\frac{\bv(\bx_i, \bz_i; \theta_t, \eta_t) \log r_{\eta}(\bz_i \g \bx_i)}{\mathbb{E}_{s(\bz_i)}\left(\bv\left(\bx_i, \bz_i; \theta_t, \eta_t\right)\right)} \!
	\right]\!.
\end{align}
where $\bv(\bx_i, \bz_i; \theta_t, \eta_t) = \frac{p_{\theta_t}(\bz_i , \bx_i)}{s(\bz_i)}$. Here $s(\bz_i)$ is a hyperproposal that has no free parameters. (We will describe it shortly.) The hyperobjective in Equation\nobreakspace \textup {(\ref {eq:rem})} is still intractable due to the expectations. We approximate it using Monte Carlo by drawing $K$ samples $\bz_i^{(1)}, \dots, \bz_i^{(K)}$ from $s(\bz_i)$. Then 
\begin{align}\label{eq:rem_2}
	\bbeta_{it}^{k} &= \frac{\bv\left(\bx_i, \bz_i^{(k)}; \theta_t, \eta_t\right)}{\sum_{k'=1}^{K} \bv\left(\bx_i, \bz_i^{(k')}; \theta_t, \eta_t\right)}
	\quad \text{and} \quad
	\mathcal{L}_{\gls{REM}}(\eta) = -\sum_{i=1}^{N} \sum_{k=1}^{K} \bbeta_{it}^{k}  \cdot \log r_{\eta}(\bz_i^{(k)} \g \bx_i), 
\end{align}
We choose the proposal $s(\bz_i)$ to be a full Gaussian whose parameters are found by matching the moments of the true posterior $p_{\theta_t}(\bz_i \g \bx_i)$. More specifically, $s(\bz_i) = \mathcal{N}(\bmu_{it}, \Sigma_{it})$ where 
\begin{align}\label{eq:moment_matching}
	\bmu_{it} &= \mathbb{E}_{p_{\theta_t}(\bz_i \g \bx_i)}[\bz_i]  \quad \text{and} \quad \Sigma_{it} =  \mathbb{E}_{p_{\theta_t}(\bz_i \g \bx_i)}\left[\left(\bz_i - \bmu_i^{(t)}\right)\left(\bz_i - \bmu_i^{(t)}\right)^\top\right]
	.
\end{align}

\begin{table*}[t]
	\centering
	\small
	\captionof{table}{Comparing \gls{REM} against the \gls{VAE} and the \gls{IWAE}. \gls{REM} uses a rich distribution $s(\bz)$ found by moment matching to learn the generative model and/or the recognition network $r_{\eta}(\bz \g \bx)$.}
	\begin{tabular}{ccccc}
	\toprule
	 Method & Objective & Proposal & Hyperobjective & Hyperproposal\\
	 \hline
	 \gls{VAE} & \acrshort{VI} & $r_{\eta}(\bz \g \bx)$ & $\text{KL}(r_{\eta}(\bz \g \bx) \vert\vert p_{\theta}(\bz \g \bx))$ & $r_{\eta}(\bz \g \bx)$\\
	 \gls{IWAE} & \acrshort{EM} & $r_{\eta}(\bz \g \bx)$ & $\mathcal{L}_{\gls{IWAE}}(\eta)$ & $r_{\eta}(\bz \g \bx)$\\
	 \gls{REM}(v1) & \acrshort{EM} & $r_{\eta}(\bz \g \bx)$ & $\text{KL}(p_{\theta}(\bz \g \bx) \vert\vert r_{\eta}(\bz \g \bx))$ & $s(\bz)$\\
	 \gls{REM} (v2) & \acrshort{EM} & $s(\bz)$ & $\text{KL}(p_{\theta}(\bz \g \bx) \vert\vert r_{\eta}(\bz \g \bx))$ &  $r_{\eta}(\bz \g \bx)$\\
	\bottomrule
	\end{tabular}
	\label{tab:approaches}
\end{table*}

The expressions for the mean and covariance matrix are still intractable. We estimate them using self-normalized importance sampling, with proposal $r_{\eta_t}(\bz_i\g \bx_i)$, and Monte Carlo. We first write
\begin{align}\label{eq:moment_matching_2}
	\bmu_{it} &= \mathbb{E}_{r_{\eta_t}(\bz_i \g \bx_i)}\left(
		\frac{\bw(\bx_i, \bz_i; \theta_t, \eta_t)}{\mathbb{E}_{r_{\eta_t}(\bz_i \g \bx_i)}\left(\bw(\bx_i, \bz_i; \theta_t, \eta_t)\right)} \bz_i
	\right),
\end{align}
(the covariance $\bSigma_{it}$ is analogous), and then estimate the expectations using Monte Carlo,
\begin{align}\label{eq:moment_matching_3}
	\bmu_{it} &\approx \sum_{k=1}^{K} \balpha_{it}^{k}\cdot \bz_i^{(k)} \text{ and } 
	\bSigma_{it} \approx \sum_{k=1}^{K} \balpha_{it}^{k} \left[(\bz_i^{(k)} - \bmu_{it})(\bz_i^{(k)} - \bmu_{it} )^\top\right]
	.
\end{align}
Note Equation\nobreakspace \textup {(\ref {eq:moment_matching_3})} imposes the implicit constraint that the number of particles $K$ be greater than the square of the dimensionality of the latents for the covariance matrix $\bSigma_{it}$ to have full rank. We lift this constraint by adding a constant $\epsilon$ to the diagonal of $\bSigma_{it}$ and setting 
\begin{align}\label{eq:moment_matching_3}
	\bSigma_{it} \approx \sum_{k=1}^{K} (\bz_i^{(k)} - \bmu_{it})(\bz_i^{(k)} - \bmu_{it} )^\top
	.
\end{align}

Algorithm\nobreakspace \ref {alg:rem} summarizes the procedure for fitting deep generative models with \gls{REM} where $\bv^{(k)}$ is computed the same way as $\bv(\bx_i, \bz_i; \theta_t, \eta_t)$. We call this algorithm \gls{REM} (v1). 

To further illustrate how \gls{REM} (v1) improves upon the \gls{IWAE}, consider replacing $s(\bz_i)$ in the definition of $\bv(\bx_i, \bz_i; \theta_t, \eta_t)$ with $r_{\eta_t}(\bz_i \g \bx_i)$. Then taking gradients of Equation\nobreakspace \textup {(\ref {eq:rem_2})} with respect to $\eta$ reduces to the \gls{IWAE} gradient for updating the recognition network $r_{\eta}(\bz_i \g \bx_i)$. Instead of using $r_{\eta_t}(\bz_i \g \bx_i)$, \gls{REM} (v1) uses a more expressive distribution found via moment matching to update the recognition network. This further has the advantage of decoupling the generative model and the recognition network as they do not use the same objective for learning. 

\parhead{Moment matching as a proposal}. We now consider using the rich moment matched distribution $s(\bz)$ to update the generative model. This changes the objective $\mathcal{L}(\theta)$ in Equation\nobreakspace \textup {(\ref {eq:rem_samples})} to 
\begin{align}\label{eq:rem_samples_new}
	\mathcal{L}(\theta) 
	&= \sum_{i=1}^{N} \sum_{k=1}^{K} \bbeta_{it}^{k}\cdot \log p_{\theta}(\bx_i, \bz_i^{(k)})
\end{align}
where $\bz_i^{(1)}, \dots, \bz_i^{(K)} \sim s(\bz_i)$ and $\bbeta_{it}^{k}$ is as defined in Equation\nobreakspace \textup {(\ref {eq:rem_2})}. We let the recognition network $r_{\eta_t}(\bz_i \g \bx_i)$ be learned the same way as done for \gls{REM} (v1). Algorithm\nobreakspace \ref {alg:rem_2} summarizes the procedure for fitting deep generative models with \gls{REM} (v2). Table\nobreakspace \ref {tab:approaches} highlights the differences between the \gls{VAE}, the \gls{IWAE}, \gls{REM} (v1), and \gls{REM} (v2).

\begin{algorithm}[t]
  \small
  \SetAlgoNoLine
     \DontPrintSemicolon
     \SetKwInOut{KwInput}{input}
     \SetKwInOut{KwOutput}{output}
     \KwInput{Data $\bx$}
     Initialize model and proposal parameters $\theta, \eta$\;
     \For{\emph{iteration} $t=1,2,\ldots$}{
       Draw minibatch of observations $\{\bx_n\}_{n=1}^{B}$\;
        \For{\emph{observation} $n=1,2,\ldots, B$}{
       Draw $\bz^{(1)}_n, \dots, \bz^{(K)}_n \sim r_{\eta_t}(\bz_n^{(k)} \g \bx_n)$ \; 
       Compute importance weights $\bw^{(k)} = \frac{p_{\theta_t}(\bz_n^{(k)} , \bx_n)}{r_{\eta_t}(\bz_n^{(k)} \g \bx_n)}$\;        Compute $\bmu_{nt} = \sum_{k=1}^{K} \frac{\bw^{(k)}}{\sum_{k=1}^{K} \bw^{(k)}} \bz^{(k)}_n$ and $\bSigma_{nt} = \sum_{k=1}^{K} \frac{\bw^{(k)}}{\sum_{k=1}^{K} \bw^{(k)}}  (\bz^{(k)}_n - \bmu_{nt})(\bz^{(k)}_n  - \bmu_n)^\top$\;
       Set proposal $s(\bz_n^{(t)}) = \mathcal{N}(\bmu_{nt}, \bSigma_{nt})$\;
       }
       Compute $\nabla_{\eta}\mathcal{L}(\eta) = \frac{1}{\vert B\vert}\sum_{n \in B}^{} \sum_{k=1}^{K} \frac{\bv^{(k)}}{\sum_{k=1}^{K} \bv^{(k)}} \nabla_{\eta}\log r_{\eta}(\bz_n^{(k)} \g \bx_n)$ and update $\eta$ using Adam\;
       Compute $\nabla_{\theta}\mathcal{L}(\btheta)  = \frac{1}{\vert B\vert}\sum_{n \in B}^{} \sum_{k=1}^{K} \frac{\bv^{(k)}}{\sum_{k=1}^{K} \bv^{(k)}} \nabla_{\theta}\log p_{\theta}(\bx_n, \bz^{(k)}_n)$ and update $\theta$ using Adam\;
      }
     \caption{Learning with reweighted expectation maximization (\acrshort{REM} (v2))\label{alg:rem_2}}
     \vspace{-2pt}
\end{algorithm}

\section{Empirical Study}\label{sec:empirical}
We consider density estimation on several benchmark datasets and compare \gls{REM} against the \gls{VAE} and the \gls{IWAE}. We find that \gls{REM} leads to significantly better performance as measured by log-likelihood on all the datasets.

\begin{figure*}[t]
	\centering
	\centerline{\includegraphics[width=1.2\textwidth, height=11cm]{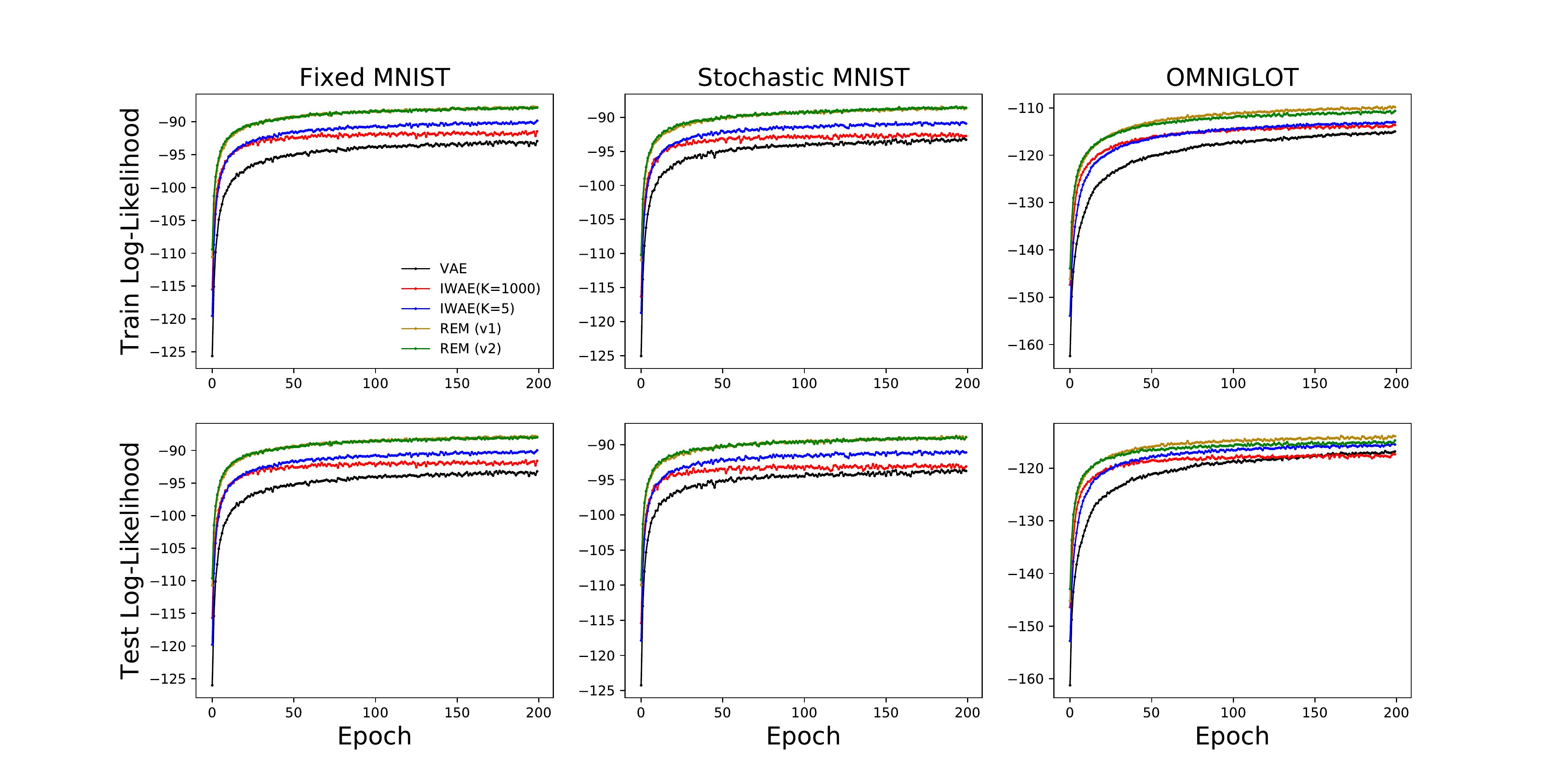}}
	\caption{\gls{REM} achieves significantly better performance than the \gls{VAE} and the \gls{IWAE} on three benchmark datasets in terms of log-likelihood (the higher the better).}
	\label{fig:nll}
\end{figure*}

\subsection{Datasets} We evaluated all methods on the \textsc{omniglot} dataset and two versions of \textsc{mnist}. The \textsc{omniglot} is a dataset of handwritten characters in a total of $50$ different alphabets~\citep{lake2013one}. Each of the characters is a single-channel image with dimension $28 \times 28$. There are in total $24{,}345$ images in the training set and $8{,}070$ images in the test set. \textsc{mnist} is a dataset of images of handwritten digits introduced by \citet{lecun1998gradient}. The first version of \textsc{mnist} we consider is the fixed binarization of the \textsc{mnist} dataset used by \citet{larochelle2011neural}. The second version of \textsc{mnist} corresponds to  random binarization; a random binary sample of digits is newly created during optimization to get a minibatch of data. In both cases the images are single-channel and have dimension $28 \times 28$. There are $60{,}000$ images in the training set and $10{,}000$ images in the test set. All these datasets are available online at \url{https://github.com/yburda/iwae}.

\subsection{Settings} We used the same network architecture for all methods. We followed~\citet{burda2015importance} and set the generative model, also called a \emph{decoder}, to be a fully connected feed-forward neural network with two layers where each layer has $200$ hidden units. We set the recognition network, also called an \emph{encoder}, to be a fully connected feed-forward neural network with two layers and $200$ hidden units in each layer. We use two additional linear maps to get the mean and the log-variance for the distribution $r_{\eta}(\bz \g \bx)$. The actual variance is obtained by exponentiating the log-variance.

We used a minibatch size of $20$ and set the learning rate following the schedule describes in \citet{burda2015importance} with an initial learning rate of $10^{-3}$. We use this same learning rate schedule for both the learning of the generative model and the recognition network. We set the dimension of the latents used as input to the generative model to $20$. We set the seed to $2019$ for reproducibility. We set the number of particles $K$ to $1{,}000$ for both training and testing. We ran all methods for $200$ epochs. 
We used Amazon EC-2 P3 GPUs for all our experiments. 

\subsection{Results}
We now describe the results in terms of quality of the learned generative model and proposal. 

\parhead{\gls{EM}-based methods learn better generative models.} We assess the quality of the fitted generative model for each method using log-likelihood. 
We report log-likelihood on both the training set and the test set. Figure\nobreakspace \ref {fig:nll} illustrates the results. The \gls{VAE} performs the worse on all datasets and on both the training and the test set. The \gls{IWAE} performs better than the \gls{VAE} as it optimizes a better objective function to train its generative model.  
Finally, both versions of \gls{REM} significantly outperform the \gls{IWAE} on all cases. This is evidence of the effectiveness of \gls{EM} as a good alternative for learning deep generative models. 

\parhead{Recognition networks are good proposals.} Here we study the effect of the proposal on the performance of \gls{REM}. 
We report the log-likelihood on both the train and the test set in Table\nobreakspace \ref {tab:rem_vs_rem}. As shown in Table\nobreakspace \ref {tab:rem_vs_rem}, using the richer distribution $s(\bz)$ does not always lead to improved performance. These results suggest that recognition networks are good proposals for updating model parameters in deep generative models. 

\begin{table*}[t]
	\centering
	\small
	\captionof{table}{\gls{REM} (v1) outperforms \gls{REM} (v2) on all but one dataset. This suggests that recognition networks are effective proposals for the purpose of learning the generative model.}
	\begin{tabular}{|cc|cc|cc|cc|}
	\cline{1-8}
	 \multicolumn{2}{|c|}{\gls{REM}} & \multicolumn{2}{c|}{Fixed MNIST} & \multicolumn{2}{c|}{Stochastic MNIST} & \multicolumn{2}{c|}{Omniglot}  \\
	 \hline
	 Proposal & Hyperproposal & Train & Test & Train & Test & Train & Test \\
	 \hline
	 $r_{\eta}(\bz \g \bx)$ & $s(\bz)$ & $\textbf{87.77}$ & $\textbf{87.91}$ & $88.68$ & $88.95$ & $\textbf{109.84}$ & $\textbf{113.94}$\\
	 $s(\bz)$ & $r_{\eta}(\bz \g \bx)$ & $87.84$ & $87.99$ & $\textbf{88.58}$ & $\textbf{88.92}$ & $110.63$ & $114.73$ \\
	\hline
	\end{tabular}
	\label{tab:rem_vs_rem}
\end{table*}
\begin{figure*}[t]
	\centering
 	\vspace*{-8pt}
	\centerline{\includegraphics[width=1.2\textwidth, height=10cm]{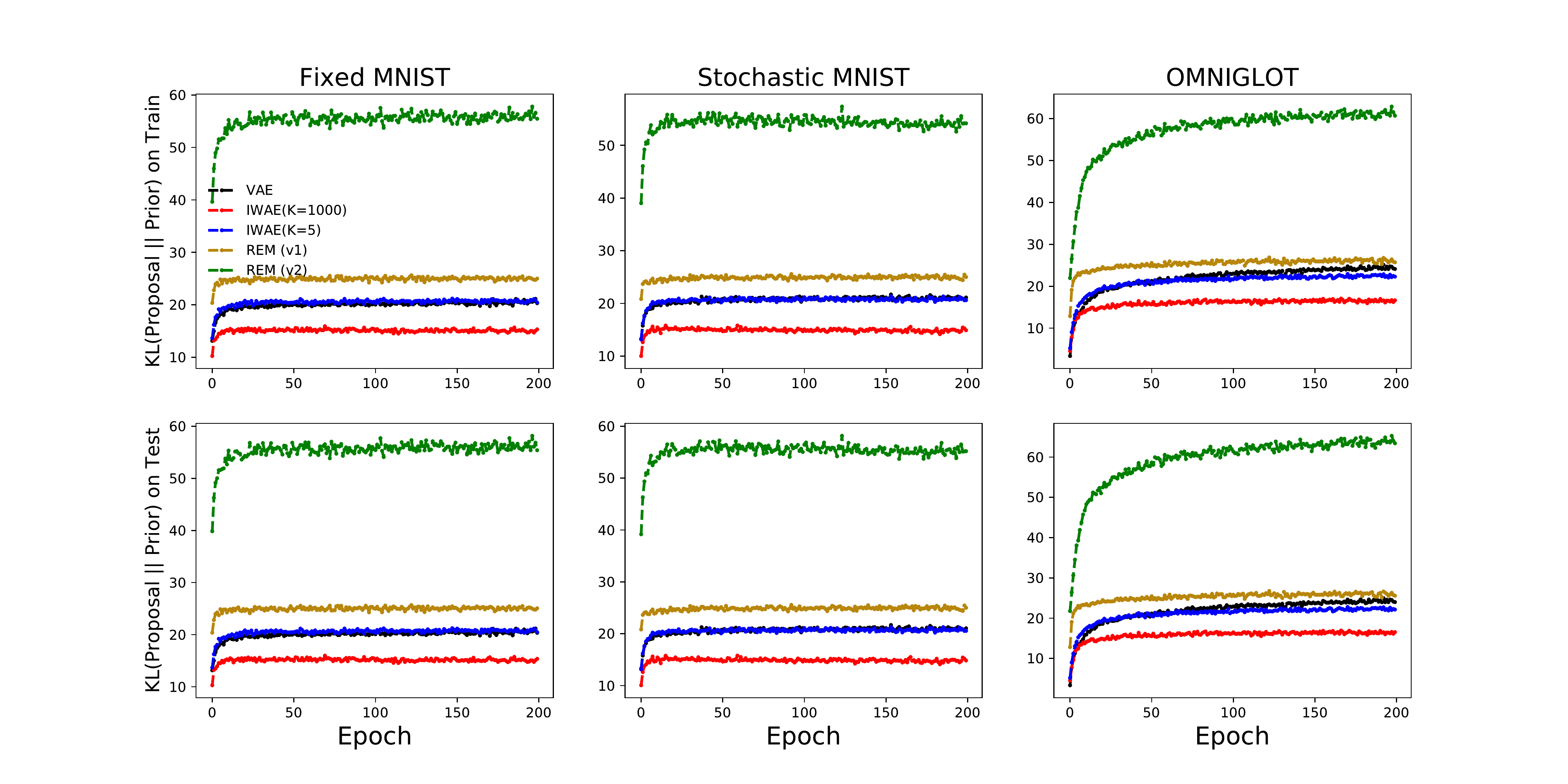}}
 	\vspace*{-4pt}
	\caption{\gls{REM} learns a better proposal than the \gls{VAE} and the \gls{IWAE}. This figure also shows that the quality of the \gls{IWAE}'s fitted posterior deteriorates as $K$ increases.}
	\label{fig:kl_prior}
\end{figure*}

\parhead{The inclusive KL is a better hyperobjective.}  We also assessed the quality of the learned proposal for each method. We use the \gls{KL} from the fitted proposal to the prior as a quality measure. This form of \gls{KL} is often used to assess latent variable collapse. Figure\nobreakspace \ref {fig:kl_prior} shows \gls{REM} learns better proposals than both the \gls{IWAE} and the \gls{VAE}. It also confirms the quality of the \gls{IWAE} degrades when the number of particles $K$ increases.

 \glsresetall

\section{Discussion}\label{sec:discussion}
We considered \gls{EM} as an alternative to \gls{VI} for fitting deep generative models. We rediscovered the \gls{IWAE} as an instance of \gls{EM} and proposed a better algorithm for fitting deep generative models called \gls{REM}. \gls{REM} decouples the learning dynamics of the generative model and the recognition network using a rich distribution found by moment matching. This avoids co-adaptation between the generative model and the recognition network. In several density estimation benchmarks, we found \gls{REM} significantly outperforms the \gls{VAE} and the \gls{IWAE} in terms of log-likelihood. 
Our results suggest we should reconsider \gls{VI} as the method of choice for fitting deep generative models. In this paper, we have shown \gls{EM} is a good alternative. 

Future work includes applying the moment matching technique used in \gls{REM} to improve variational sequential Monte Carlo techniques~\citep{naesseth2017variational, maddison2017filtering, le2017auto} or using \gls{REM} together with doubly-reparameterized gradients~\citep{tucker2018doubly} to fit discrete latent variable models.

\parhead{Acknowledgements. } We thank Scott Linderman, Jackson Loper, and Francisco Ruiz for their comments. ABD is supported by a Google PhD Fellowship.

\bibliographystyle{apa}
\bibliography{main,refs}

\end{document}